\documentclass[runningheads,letterpaper]{llncs}

\usepackage{amssymb}
\usepackage{graphicx}
\usepackage{amsmath}
\usepackage{textcomp}
\usepackage{stmaryrd}
\usepackage{rotating}
\usepackage{mdframed}

\usepackage{url}
\urldef{\mailsa}\path|r.k.moore@sheffield.ac.uk|    
\newcommand{\keywords}[1]{\par\addvspace\baselineskip
\noindent\keywordname\enspace\ignorespaces#1}

\begin{document}

\mainmatter

\title{PCT and Beyond:\\Towards a Computational Framework for `Intelligent' Communicative Systems}

\titlerunning{PCT and Beyond}

\author{Roger K. Moore}

\authorrunning{R. K. Moore: PCT and Beyond}

\institute{Department of Computer Science, University of Sheffield\\
Regent Court, 211 Portobello, Sheffield, S1 4DP, UK\\
\mailsa\\
\url{http://www.dcs.shef.ac.uk/~roger}}

\toctitle{PCT and Beyond: Towards a Computational Framework for `Intelligent' Communicative Systems}
\tocauthor{Roger K. Moore}
\maketitle

\begin{abstract}

Recent years have witnessed increasing interest in the potential benefits of `intelligent' autonomous machines such as robots.  Honda's \emph{Asimo} humanoid robot, iRobot's \emph{Roomba} robot vacuum cleaner and Google's driverless cars have fired the imagination of the general public, and social media buzz with speculation about a utopian world of helpful robot assistants or the coming robot apocalypse!  However, there is a long way to go before autonomous systems reach the level of capabilities required for even the simplest of tasks involving human-robot interaction - especially if it involves communicative behaviour such as speech and language.  Of course the field of Artificial Intelligence (AI) has made great strides in these areas, and has moved on from abstract high-level rule-based paradigms to embodied architectures whose operations are grounded in real physical environments.  What is still missing, however, is an overarching theory of intelligent \emph{communicative} behaviour that informs system-level design decisions in order to provide a more coherent approach to system integration.  This chapter introduces the beginnings of such a framework inspired by the principles of Perceptual Control Theory (PCT).  In particular, it is observed that PCT has hitherto tended to view perceptual processes as a relatively straightforward series of transformations from sensation to perception, and has overlooked the potential of powerful generative model-based solutions that have emerged in practical fields such as visual or auditory scene analysis.  Starting from first principles, a sequence of arguments is presented which not only shows how these ideas might be integrated into PCT, but which also extend PCT towards a remarkably symmetric architecture for a needs-driven communicative agent.  It is concluded that, if behaviour is the \emph{control} of perception (the central tenet of PCT), then perception (at least for communicative agents) is the \emph{simulation} of behaviour.


\end{abstract}

\section{Introduction} \label{sec:INTRO}

The past few decades have seen an enormous growth in the level of interest being shown in so-called `intelligent' machines \cite{Kurzweil1990}.  Funding agencies and large corporations worldwide have been investing heavily in autonomous systems - especially \emph{robotics} - on the premise that significant economic benefits may be derived from automating hitherto people-intensive activities.  Ranging from production-line robots to homecare assistants, and from robotic surgeons to driverless cars, future intelligent systems are expected to transform our lives in much the same way that the invention of the steam engine accelerated the industrial revolution during the 18\textsuperscript{th} and 19\textsuperscript{th} centuries \cite{Roy2000a,Kanda2004,Gockley2005,Levy2007,Diehl2012}.

However, in order to deliver the expected benefits, future autonomous systems will indeed need to be intelligent - they must integrate seamlessly into real-world environments, act appropriately in complex physical and temporal situations, solve difficult logistical problems, interact effectively with human users/operators using accepted social conventions (such as speech and language), be robust in the face of unpredictable disturbances and interruptions, operate independently within an accepted ethical framework, and be at least partially responsible for their own behaviours \cite{Winfield2012}.  This is a challenging wish list that goes well beyond the bounds of traditional fields such as Artificial Intelligence (AI).  Indeed, the requirements for intelligent systems/robots are so demanding that insights need to be integrated from a wide array of disciplines ranging from engineering and computer science to psychology, cognitive neuroscience and linguistics.

In practice, a robot is a complex physical entity often consisting of a large number of moving parts, an electrical power system, an array of electronic components including on-board and off-board processors, wired and wireless communication links, various computer operating systems and a range of software modules for managing the overall system.  This means that it is not only necessary to develop comprehensive tools and techniques for designing, building and programming such devices to meet particular application requirements, but it is very likely that approaches will also need to be based on a deeper understanding of how existing intelligent systems - living organisms - solve the challenges listed above \cite{Delcomyn2007,Floreano2008}.

At present, the sheer complexity of such systems, coupled with the high cost of developing bespoke hardware platforms, inevitably means that components are integrated without a great deal of thought being given to more general principles of intelligent behaviour.  Off-the-shelf solutions for locomotion, navigation, manipulation and interaction are often combined with independent modules for input/output modalities such as vision, speech and gesture without too much consideration of potential synergies.  As a result, behaviours may be programmed in an ad-hoc manner using heuristic (rather than principled) approaches to the necessary algorithms\footnote{Of course it's not that such theory doesn't exist.  Rather, the demands of contemporary intelligent systems are such that it is often necessary to take a pragmatic approach to system implementation.}.  What is required is an overarching theory of intelligent behaviour that informs system-level design decisions in order to provide a more coherent approach to system integration, especially for communicative agents.

\subsection{Good Old-Fashioned Artificial Intelligence} \label{sec:GOFAI}

The term Artificial Intelligence (AI) was coined in 1955 and, in its early years, was mainly concerned with mathematical logic and automatic theorem proving (on the assumption that `intelligence' was founded on processes of rational thought) \cite{Russell2003}.  Activities such as playing chess were seen as the epitome of human intellectual achievement, and thus became the focus of early research. However, it was soon realised that physically manipulating actual chess pieces could be more challenging than playing the game itself.  AI thus moved on to fulfil an important role in expanding our understanding of how living systems in general, and human beings in particular, interact physically with the world through developments in areas such as bio-inspired robotics and autonomous systems.

\begin{figure}[!h]
\centering
\includegraphics[width=12cm]{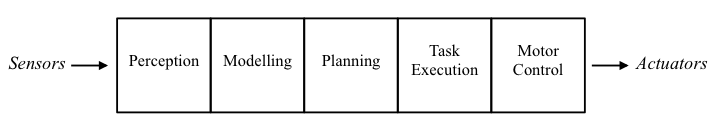}
\caption{Illustration of the sequential deliberative (\emph{Sense} \textrightarrow\;\emph{Plan} \textrightarrow\;\emph{Act}) architecture popular in Good Old-Fashioned Artificial Intelligence (GOFAI).}
\label{fig:SPA}
\end{figure}

Until the mid-1980s, the main paradigm for AI-based robotics was the so-called `deliberative' architecture in which symbolic representations of the world were manipulated in a hierarchical rule-based framework involving goals and subgoals.  The process operated using a \emph{Sense} \textrightarrow\;\emph{Plan} \textrightarrow\;\emph{Act} cycle (see Fig.~\ref{fig:SPA}), very much in tune with the `behaviourist' \emph{Stimulus} \textrightarrow\;\emph{Response} (S-R) framework that was dominating the field of psychology at the time \cite{Brunswik1956}.

When applied to robotics, the deliberative AI paradigm essentially took a high-level approach in which functions such as navigation were regarded as problem-solving activities that required the manipulation and search of appropriate symbolic data structures.  However, it soon became apparent that this perspective - subsequently termed Good Old-Fashioned Artificial Intelligence (GOFAI)  \cite{Haugeland1985} - suffered from severe limitations, particularly with respect to an over-reliance on accurate world models and an inability to respond quickly to changing situations and context.

\subsection{Behaviour-Based Robotics} \label{sec:BBR}

\begin{figure}[!h]
\centering
\includegraphics[width=12cm]{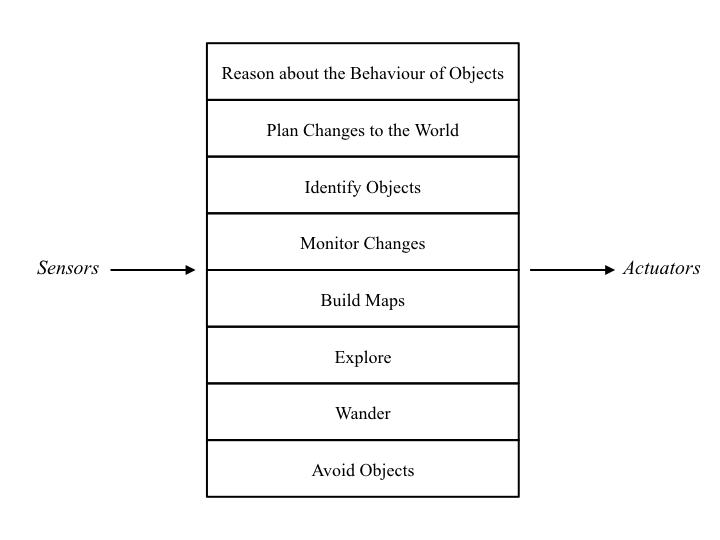}
\caption{Illustration of the layered reactive subsumption architecture popular in behaviour-based robotics.}
\label{fig:SUB}
\end{figure}

By the 1980s, the mounting difficulties faced by attempting to deploy GOFAI in real-world robots had led to a new `reactive' (as opposed to `deliberative') approach based on low-level \emph{representation-free} processes.  Hailing the establishment of what was to be known as `New AI', Rodney Brooks introduced a novel `subsumption' architecture in which the emphasis was on real-time behaviour using simple computations embedded in a layered structure \cite{Brooks1986}, \cite{Brooks1991} (see Fig.~\ref{fig:SUB}).  The basic idea was that the different layers should operate more or less independently (and in parallel), with the higher levels relying on successful operation of the lower levels.  If necessary, the higher levels could change (that is, `subsume') the behaviour of the lower levels.  Overall, the emphasis was on the \emph{grounding} of behaviour in the real-world, with the hypothesis that complex interactions should arise as an \emph{emergent} property of an array of simple processes (as had been elegantly proposed by Braitenberg in the 1980s \cite{Braitenberg1984}).

The subsumption approach initiated a trend towards what became known as `behaviour-based robotics' \cite{Arkin1998}, and the emphasis shifted from high-level abstract problem solving to low-level grounded intelligence.

Behaviour-based robotics was sufficiently successful in solving practical real-world problems that, in 1990, Rodney Brooks founded the company iRobot\textsuperscript{\textregistered} to develop and market the Roomba\textsuperscript{\textregistered} vacuum cleaning robot.  More than 10 million Roomba robots have been sold worldwide, making it the most successful mass-market robot to date.  However, the subsumption approach has been criticised for being difficult to scale up to more complex robots.  Also, the emphasis on representation-free processing, whilst being popular with a subset of psychologists, is ultimately a significant restriction\footnote{In reality, the proposal that internal models were unnecessary was a stance adopted by Brooks to make a strong point about the inadequacies of GOFAI.  In practice, the subsumption architecture incorporates such models (for example, the building of maps is mentioned in Fig.~\ref{fig:SUB}).}.

\subsection{Artificial Cognitive Systems} \label{sec:ACS}

Recent work in robotics has taken a more pragmatic view by mixing and matching the best ideas from GOFAI and behaviour-based robotics with some of the new perspectives emerging from the field of cognitive neuroscience.  Known as `artificial cognitive systems' \cite{Vernon2007,Vernon2007a}, the most significant influence has been the emphasis on enaction, embodiment and situatedness in which the relationship between a goal-driven robot and its physical real-world context is paramount \cite{Vernon2010a,Vernon2010b}.  Low-level interactions are managed locally and high-level representations are given `meaning', not by hand-crafted rules, but by virtue of their grounding in real-world interactions.  Interaction is facilitated by the `affordances' \cite{Gibson1977,Geng2013} provided by a robot's environment.

\begin{figure}[!h]
\centering
\includegraphics[width=12cm]{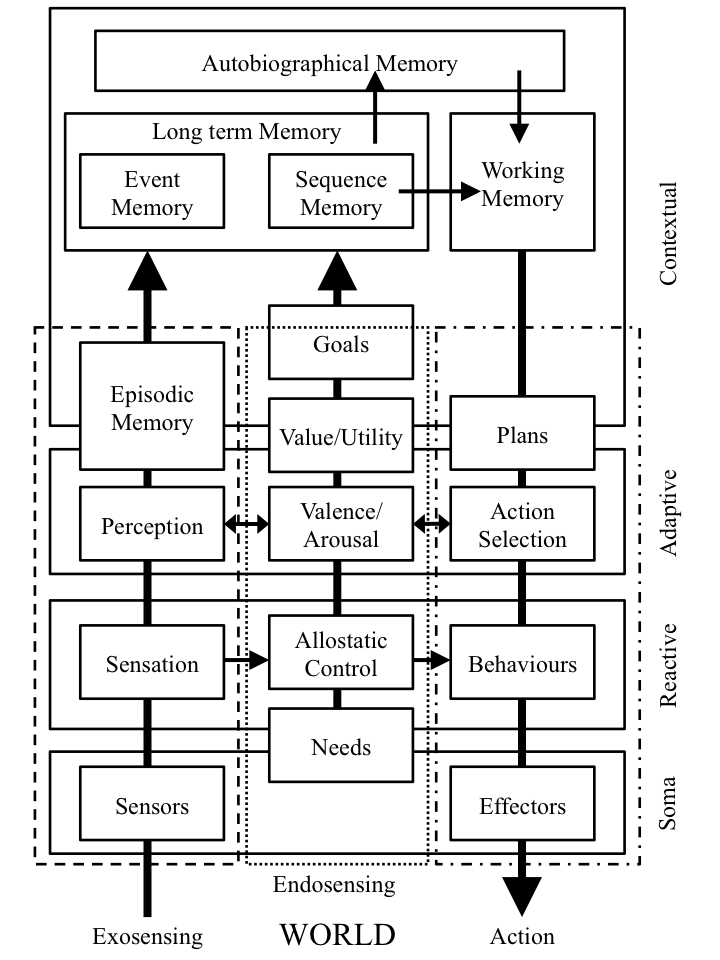}
\caption{Illustration of the Distributed Adaptive Control (DAC) architecture organised into exosensing, endosensing and action columns, and soma, reactive, adaptive and contextual layers (derived from Fig. 1 in \cite{Verschure2012}).}
\label{fig:DAC}
\end{figure}

A particular example of a contemporary architecture for an artificial cognitive system is Distributed Adaptive Control (DAC) \cite{Pfeifer1992,Verschure2012}.  The DAC architecture is based on the hypothesis that the living brain maintains a stable relationship with its environment by continuously solving the \emph{How, Why, What, Where, When} (H4W) problem.  Implementation uses an artificial neural structure organised into soma, reactive, adaptive and contextual layers, and columns which represent exosensing (defined as the sensation and perception of the world), endosensing (detecting and signalling states derived from the physical self) and action (the interface to the world) - see Fig.~\ref{fig:DAC}.

Unlike their GOFAI predecessors, modern cognitive approaches to intelligent systems (such as DAC) view action and perception as being synergistic (rather than separate) processes \cite{Verschure2003}, and that skills should be acquired through the robot's own active exploration of the world (rather then being pre-programmed by the system designer) \cite{Demiris2008,Cangelosi2015}.  Such approaches very much reflect contemporary models of living systems, particularly the discovery in the 1990s of so-called `mirror neurons' - neural structures which appear to provide a vital link between sensory and motor behaviour and thereby a mechanism for action understanding, imitation and learning \cite{Rizzolatti1996,Rizzolatti2004,Oztop2006}.

A recent book by Murray Shanahan provides an excellent overview of the contemporary perspective in artificial cognitive systems \cite{Shanahan2010}.

\subsection{Agent Based Modelling} \label{sec:ABM}

\begin{figure}[!h]
\centering
\includegraphics[width=12cm]{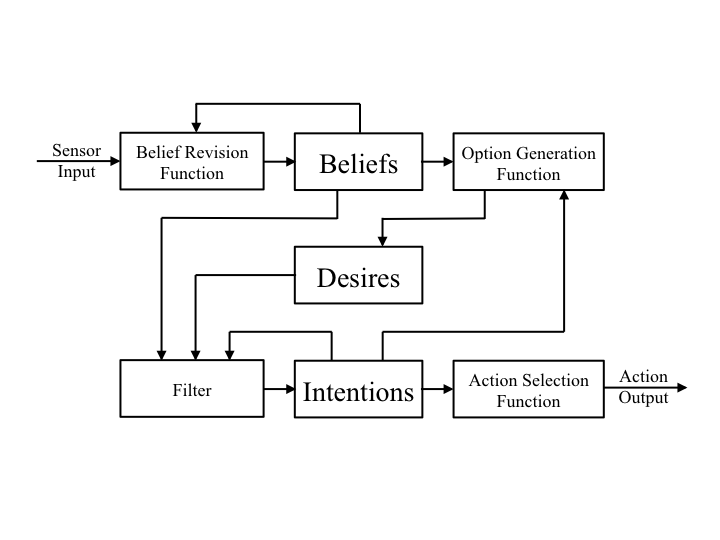}
\caption{Illustration of the Beliefs, Desires and Intentions (BDI) architecture used in the field of Agent-Based Modelling.  `Beliefs' capture the informational state of an agent, `Desires' represent its motivational state and `Intentions' represent its deliberative state.}
\label{fig:BDI}
\end{figure}

One aspect of behaviour that is common across alternative approaches to modelling and building intelligent systems is \emph{intentionality}.  It is clearly the case that living systems appear to be goal-directed and purposeful in their endeavours \cite{Csibra2007}, and this has had a major influence on AI and robotics.   In particular, one area in which intentionality plays a key role is the field of Agent-Based Modelling (ABM) \cite{Gilbert2008,Richetin2010}.  ABM is a well-established methodology for simulating the actions and interactions of multiple agents: for example, predicting the behaviour of crowds, optimising a supply chain or managing a workforce.   Various modelling paradigms are employed, such as cellular automata \cite{Wolfram1983} or dedicated multi-agent programming environments \cite{Tisue2004}.  However, for `intelligent' agents, ABM simulations are often constructed using a Beliefs Desires Intentions (BDI) architecture \cite{Rao1995,Wooldridge2000} on the premise that such internal structures are required to adequately condition the behaviour of individual agents - see Fig.~\ref{fig:BDI}.

BDI is a powerful approach to modelling agents, and it has been applied successfully to robotics \cite{Davis2010,Lincoln2013}.  However, BDI does not specify how to recognise/interpret behaviour under conditions of ambiguity or uncertainty - a crucial feature of intelligent systems.

\subsection{Contemporary Intelligent Systems} \label{sec:CIS}

Over the past ten years, the field of robotics and autonomous systems has grown in stature and achievement.  Robots such as Boston Dynamics'  \emph{Big Dog} and Honda's \emph{Asimo} have successfully demonstrated that modern-day computing and electronics are finally fast enough to permit real-time control of complex behaviours such as running or climbing on uneven surfaces.  Robotic quadcopters have been shown flying through moving hoops \cite{TheDmel2010} and playing tennis \cite{InfoLeak2011}, robot hands have been programmed to catch thrown objects \cite{Mazhar2013} and NASA's \emph{Robonaut R2} has been assisting on the International Space Station since 2011 \cite{Robonaut2015}.  Also, recent years have seen tremendous growth in the field of Artificial Neural Networks (ANNs), particularly with the success of \emph{deep learning} as a mechanism to optimise the parameters of such multilayered networks on massive amounts of training data \cite{Hinton2007,Mnih2015}.

Nevertheless, despite this tremendous progress, there is still a long way to go before artificial intelligent systems will be able to demonstrate the flexibility, robustness and autonomy exhibited by even the simplest living organism \cite{Pfeifer1999,DeKamps2012}.  Many contemporary robots rely on teleoperation by human operators to overcome weaknesses in their overall design, and attempts to create humanoid robots are fraught with difficulties ranging from the risk of repulsing human users due to the uncanny valley effect \cite{Mori1970,Moore2012}, to the more general uncertainties associated with human-robot interaction \cite{Maxwell2007,Scheutz2007,Herrmann2010} - especially if it involves communicative agents and spoken language \cite{Moore}.

For example, probably the most well known communicative agent is \emph{Siri}, Apple's voice-based personal assistant for the iPhone.  Released in 2011, \emph{Siri} suddenly brought spoken language technology to the attention of the non-specialist user.   Voice dictation software for document creation on PCs had been available since the 1990s, culminating with the release of Dragon System's \emph{Naturally Speaking} and IBM's \emph{ViaVoice} products in 1997.  The big difference was that \emph{Siri} combined automatic speech recognition and speech synthesis with natural language processing and dialogue management in order to facilitate a more \emph{conversational} interaction between users and smart devices, and competitors such as \emph{Google Now} and Microsoft's \emph{Cortana} soon followed.

\begin{figure}[!h]
\centering
\includegraphics[width=12cm]{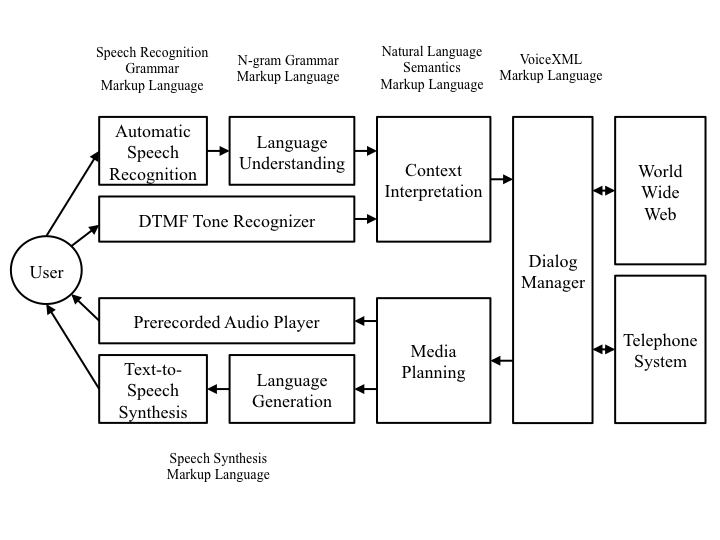}
\caption{Illustration of the standard \emph{Speech Interface Framework} published by the World Wide Web Consortium (W3C) \cite{W3C}.}
\label{fig:W3C}
\end{figure}

In reality, the practical value of contemporary communicative agents is somewhat in doubt (as evidenced by the preponderance of videos on \emph{YouTube} which depict humorous rather than practical interactions).  This has been confirmed by a recent survey that discovered only 13\% of respondants use their voice-based personal agent daily, whereas 46\% had tried it once and then abandoned it \cite{Liao2015}.  One reason for the lack of usability is that contemporary communicative agents are founded on a classic \emph{stimulus}\textrightarrow\emph{response} architecture (as illustrated in Fig.~\ref{fig:W3C}); a user speaks, their utterance is processed, a response is formulated and the system speaks back.  Such an approach completely overlooks the reality of \emph{languaging} as an emergent property of the dynamic coupling between \emph{intentional} agents that serves to facilitate distributed sense-making through cooperative behaviours \cite{Maturana1987,Cummins2004,Bickhard2007,Cowley2011,Fusaroli2014}.  Furthermore, the contemporary view is that language is based on the co-evolution of two key traits: \emph{ostensive-inferential} communication and \emph{recursive mind-reading} \cite{ScottPhillips2015}.

These are very sophisticated notions but, as yet, there is no clear practical framework for implementing such concepts.  As a result, the field is still open for new ideas, and some of them may not be so new.  For example, what might Perceptual Control Theory (PCT) - established over 50 years ago \cite{Powers1960,Powers1960a} (and the topic of this volume) - have to contribute to future `intelligent' communicative systems?

\section{Whither Perceptual Control Theory?} \label{sec:WPCT}

Although it is more than 40 years since the publication of Bill Powers' seminal book `Behavior: The Control of Perception' (B:CP) \cite{Powers1973}, Perceptual Control Theory may still have something important to say about computational models of `intelligent' communicative behaviour.  Over the years, the main thrust of PCT has been to question the traditional behaviourist \emph{stimulus}\textrightarrow\emph{response} stance that still prevails in some sections of the psychology field \cite{Bourbon1999,Mansell2011}.  PCT aims to provide a more parsimonious model which highlights the critical importance of acknowledging the existence of internal preferred states (so called `reference variables') arranged in a control hierarchy, together with the power of negative feedback to achieve and maintain those preferred states in the face of unpredictable disturbances.  Crucially, PCT focuses attention on the control of perceptual \emph{inputs} rather than behavioural \emph{outputs} \cite{Powers2008}.

Of course, negative feedback control is endemic in the field of robotics (particularly at the lowest levels of motor control), not least because many robotocists are trained control engineers.  However, mainstream control engineering tends to concentrate on the regulation of required behaviour (low-level outputs) in a single control loop, and the emphasis is on modelling the dynamics of the system under control in order to calibrate the parameters of the control process.  Also, hierarchical control structures are much less common.  PCT thus offers a broader perspective that is often overlooked, as well as introducing the proposal that living organisms control \emph{perceptions} rather than behaviours.

\subsection{Classic Automatic Control} \label{sec:CAC}

\begin{figure}[!h]
\centering
\includegraphics[width=12cm]{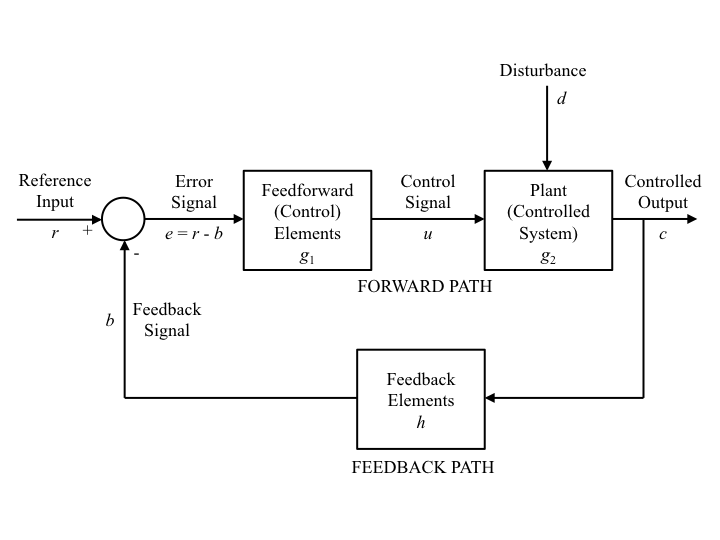}
\caption{Illustration of a classic negative feedback control system from automatic control theory (derived from Fig. 2-6 on page 16 of \cite{DiStefanoIII1990}).}
\label{fig:CCS}
\end{figure}

As a reminder, PCT was developed in the context of existing knowledge of the classic theory of automatic control - the study of systems that are capable of self-regulation, usually through the mechanism of negative feedback \cite{DiStefanoIII1990}.  A classic single-input single-output closed loop negative feedback control system from automatic control theory is illustrated in Fig.~\ref{fig:CCS}.  The plant (or controlled system) $g_{2}$ is the process controlled by the feedback control system, and the behaviour of the plant may be subject to arbitrary disturbances $d$.  The feedforward (control) elements $g_{1}$ generate control signals $u$ that are applied to the plant which produces the controlled output $c$.  The reference input $r$ specifies the desired output of the plant.  The feedback elements $h$ map the controlled output $c$ to the feedback signal $b$, and this is compared to the external reference input $r$ and the resulting error signal $e$ generates a control action via $g_{1}$.  With appropriate functions for $g_{1}$ , $g_{2}$ and $h$, the system should compensate for any disturbances and stabilise with $b \to r$ \emph{without having to measure} $d$.  The process is termed \emph{negative} feedback because the comparator involves subtraction.

It can be shown that the input-output relations of the classic control system shown in Fig.~\ref{fig:CCS} are given (in the frequency domain) by the equation
\begin{equation}
  C = \left( \frac{G_{2}G_{1}}{1 + HG_{2}G_{1}} \right) R    ,
\end{equation}\smallskip
where $\frac{G_{2}G_{1}}{1 + HG_{2}G_{1}}$ is the closed-loop transfer function and $HG_{2}G_{1}$ is the loop gain.  If $G_{2}G_{1} \gg 1$ and $H \approx 1$, then $C \approx R$.

Clearly the tracking behaviour of a classic negative-feedback control system depends on the feedforward (control) elements $g_{1}$.  If the control element is slow to respond to a disturbance, then stabilisation may take too long; this is referred to as an overdamped system.  On the other hand, if the loop gain is too high, then the system may overshoot and even oscillate; this is referred to as an underdamped system.  In practice,  $g_{1}$ is often implemented using a Proportional, Integral, Derivative (PID) controller which takes the following \emph{idealised} form:
\begin{equation}
  u_{PID} = K_{P}e + K_{I}\int e\left( t \right) dt + K_{D} \frac{de}{dt}    ,
\end{equation}
where the three constants - $K_{P}$, $K_{I}$ and $K_{D}$ - are used to optimise the stability of the control process (referred to as a critically damped system).  $K_{P}$ is known as the `loop gain' and it determines the speed of response of the system to error.  If $K_{P}$ is too high, then the system may respond too quickly and overshoot (or even become unstable); if it is too low, then the system may respond too late to counteract any disturbance.  The effect of $K_{D}$ is to slow the rate of change of the controller output and thus minimise overshoot arising from a high $K_{P}$.  $K_{I}$ is intended to reduce any residual steady-state error.  There are various methods for estimating these constants and, in practice, $K_{I}$ and $K_{D}$ can often be set to zero.

\subsection{Hierarchical Perceptual Control} \label{sec:HPCT}

\begin{figure}[!h]
\centering
\includegraphics[width=12cm]{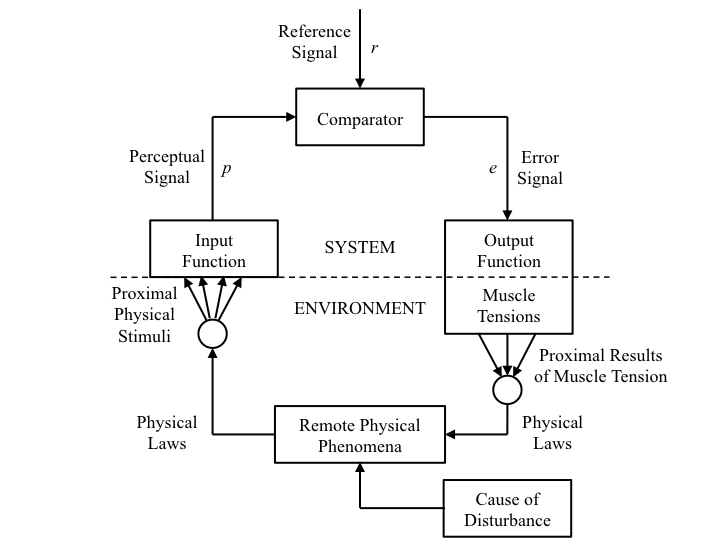}
\caption{Illustration of a basic perceptual control unit (derived from Fig. 5.2 on page 61 of \cite{Powers1973}).}
\label{fig:PCS}
\end{figure}

Contemporary automatic control systems tend to be based on a single control loop, but the dynamic behaviour of the system to be controlled is often very complex.  PCT, on the other hand, decomposes a system into a hierarchical/layered structure with a multiplicity of control loops at each level; this is referred to as \emph{Hierarchical Perceptual Control Theory} (HPCT) \cite{Powers1973}.  Inspired by classic automatic control theory, the basic perceptual control unit (feedback control loop) in PCT is illustrated in Fig.~\ref{fig:PCS}.

Note the similarity between Fig.~\ref{fig:PCS} and Fig.~\ref{fig:CCS}.  The main difference is that the PCT structure illustrated in Fig.~\ref{fig:PCS} is specifically aimed at modelling the behaviour of a \emph{living} system, and this means that the reference variable is intrinsic (rather than extrinsic).  In all other respects the two control structures are effectively identical\footnote{Note that PCT practitioners refer to the feedback signal ($p$ in Fig.~\ref{fig:PCS}) as the \emph{controlled variable} thereby emphasising that it is the perceptual signal that is controlled in a negative feedback control system, not the plant.}.

As mentioned, HPCT derives its power from the hierarchical structure of multiple control loops in which the output from one loop provides the reference signal for another loop.  Powers \cite{Powers1973} proposed specific orders of control (starting at the lowest first-order level) as follows: \emph{intensity $\Leftrightarrow$ sensation $\Leftrightarrow$ configuration $\Leftrightarrow$ transitions $\Leftrightarrow$ sequence $\Leftrightarrow$  relationships $\Leftrightarrow$ program $\Leftrightarrow$ principles $\Leftrightarrow$ system concepts}.

\begin{figure}[!h]
\centering
\includegraphics[width=12cm]{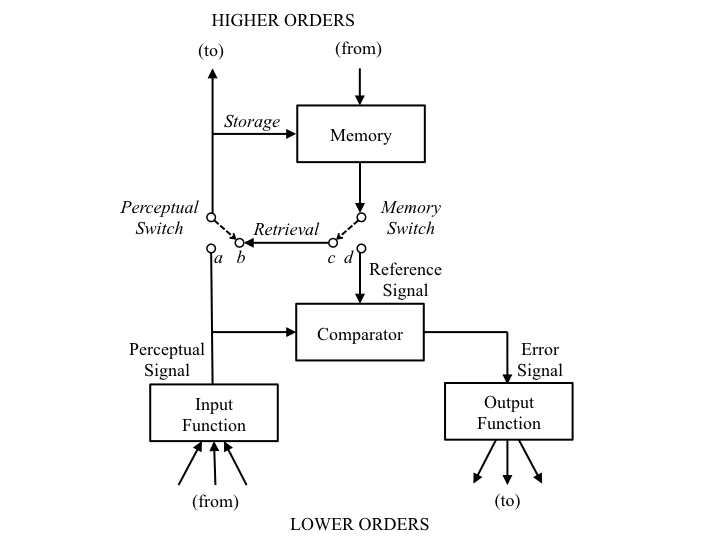}
\caption{Illustration of an HPCT unit showing the perceptual and memory switches which can be set for control mode (\emph{a+d}), passive observation mode (\emph{a+c}), automatic mode (\emph{b+d}) or imagination mode (\emph{b+c}) (derived from Fig. 15.2 on page 221 of \cite{Powers1973}).}
\label{fig:HPCTU}
\end{figure}

Powers also generalised the basic perceptual control unit to include the storage and retrieval of information in memory.  In particular, he proposed that all reference signals constitute recordings of past perceptual signals that are retrieved by address signals from the higher level - see Fig.~\ref{fig:HPCTU}.  Implementation of this feature required the addition of two switches - a memory switch and a perceptual switch.  Powers noted that each of the four possible combinations of switch settings lead to interesting outcomes.  If both switches are vertical (\emph{a+d}), then the loop is in ``\emph{conventional control mode}''.  If the perceptual switch is vertical and the memory switch is non-vertical (\emph{a+c}), then information can be acquired without action taking place; Powers called this ``\emph{passive observation mode}''.  If the memory switch is vertical and the perceptual switch is non-vertical (\emph{b+d}), then control takes place with no perceptual awareness at higher levels; Powers referred to this as ``\emph{automatic mode}''.   Finally, if both switches are non-vertical (\emph{b+c}), then past perceptions (retrieved from memory) are redirected back up the hierarchy; Powers designated this as ``\emph{imagination mode}'', and identified it as a potential mechanism for visualisation and planning.

On the surface, the HPCT architecture looks remarkably similar to subsumption (as described in section \ref{sec:BBR}).  However, a key distinction is that HPCT is capable of optimising many variables simultaneously within a negative feedback control framework, whereas subsumption is essentially an event-driven \emph{stimulus}\textrightarrow\emph{response} architecture \cite{Dalchow2012}.

Finally, Powers identified three types of learning within HPCT: the storage of information in memory, problem-solving and \emph{reorganisation}.  The latter is particularly interesting since it involves altering the structure and properties of the control systems themselves.  Powers viewed reorganisation as a kind of meta control system that had the objective of reducing the intrinsic error of the overall system to zero.

\subsection{The Way Forward?} \label{sec:TWF}

Perceptual control theory (and HPCT) appears to offer a number of interesting features for modelling the behaviour of living systems \cite{Taylor1999}, and should thus be a serious contender for implementing control in artificial systems such as robots (as envisaged by Powers \cite{Powers1979,Powers1979a,Powers1979b,Powers1979c}).  A few attempts have been made in this area by PCT practitioners \cite{Kennaway1999,Kennaway2004,YoungURL}.  However, since a large proportion of robotocists are control engineers who take negative feedback systems as a given, they have not seen the particular benefits of using (H)PCT.

On the other hand, (H)PCT has started to have some influence in the field of spoken language processing \cite{Moore2007a}.  It was established many years ago that human speech is optimised \emph{by} speakers \emph{for} listeners \cite{Lombard1911}, and this has been theorised as resulting from the operation of negative feedback processes \cite{Lindblom1990}.  Hence, there is an obvious link with PCT.  However, mainstream speech technology systems have tended to ignore speaker-listener dependencies and such behaviour is simply modelled as stochastic variation \cite{Pieraccini2012}.

In an attempt to capture such dependencies, and inspired by classic automatic control, (H)PCT and recent discoveries in cognitive neuroscience (such as mirror neurons - discussed in Section \ref{sec:ACS}), the PREdictive SENsory Control and Emulation (PRESENCE) architecture \cite{Moore2007b} provides some novel technical solutions in this area.  For example, a PRESENCE-based text-to-speech synthesiser - \emph{C2H} - has been developed that is capable of adjusting its pronunciation while it is speaking as a function of its perceived communicative success, the latter being judged using a perceptual feedback path in which the `controlled variable' was an estimate of the intelligibility \cite{Moore2011,Nicolao2012}.

\begin{figure}[!h]
\centering
\includegraphics[width=12cm]{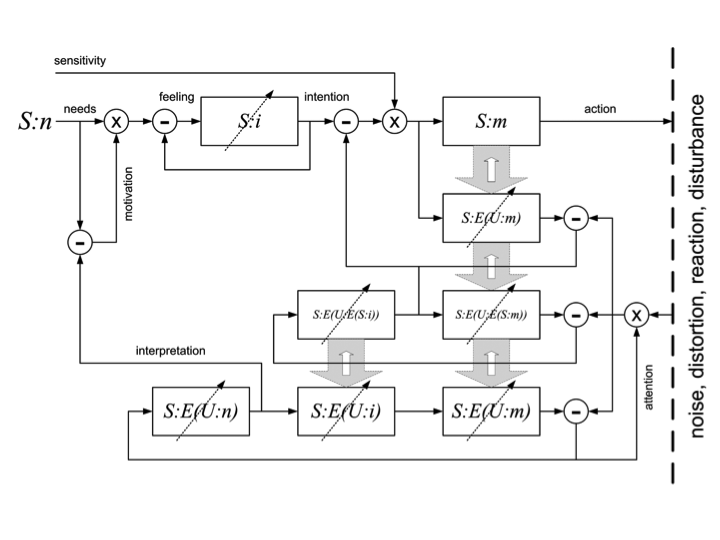}
\caption{Illustration of the PCT-inspired PREdictive SENsorimotor Control and Emulation (PRESENCE) architecture for spoken language processing \cite{Moore2007a,Moore2007b}).  The diagram depicts a high-level view of an integrated speaker-listener; the first level is the primary route for motor behaviour (speaking), the second layer is an emulation of possible motor actions, the third layer is an emulation of the interlocutor's emulation of the speaker-listener and the bottom layer is the interpretation of the interlocutor (listening).  All levels involve control feedback processes, and information is shared between one layer and another.  The architecture is intended to describe a system that can communicate successfully in the face of disturbances such as noise, distortion and feedback from the interlocutor.}
\label{fig:PRES}
\end{figure}

PRESENCE not only provides a novel architecture for generating spoken language, it also shows how the principles of feedback control can be applied to the recognition and interpretation of spoke language \cite{Moore2010}.  Indeed, PRESENCE emphasises that in order to manage its own behaviour (its perceptions), a social agent also needs to be able to interpret the world and the behaviour of other intelligent systems within it.  Hence, much greater emphasis is placed on modelling perception than is the norm in mainstream PCT.  PRESENCE thus provides an interesting insight into a potentially more general-purpose PCT-inspired architecture that is not specific to spoken language processing and which could be applied in engineered solutions.  However, the founding principles of such an architecture have not been fully elucidated, so what follows is a step in that direction.

\section{Towards `Intelligent' Communicative Systems} \label{sec:TIS}

The insightful catchphrase for (H)PCT is \emph{\textquotedblleft Behaviour is the control of perception\textquotedblright} - but what is perception, and how are controlled variables estimated?  In B:CP \cite{Powers1973}, Powers himself does not say a great deal about perceptual mechanisms; he acknowledges the potentially arbitrary relationship that exists between internal perceptions and external reality, but he seems to assume some form of straightforward (neural) transformation/mapping from first-order intensities to \emph{invariant} higher-order perceptions that is carried out by an `input function' (albeit mediated by memory - see Section \ref{sec:HPCT}).  Of course such an assumption is common in the pattern recognition/machine learning literature.  However, in practical fields such as visual or auditory scene analysis (and even in the latest theories of brain function \cite{Friston2009}), the notion of `perception as transformation' has been complemented by the emergence of more powerful (and more successful) \emph{generative model-based} solutions \cite{Weiss2001,Radfar2008,Gales2007}.  Hence, there is a potentially valuable opportunity to map these findings back into (H)PCT.

Not only is this an interesting route to take, but it also turns out to be a crucial step towards a deeper understanding of how one intelligent system (such as an autonomous social robot) might interact/communicate with another (such as a human being) \cite{Moore2014}.  What follows, therefore, is a fundamental analysis - starting from first principles - that intersects with (H)PCT and then extends it in important and interesting ways.

\subsection{Actions and Consequences} \label{sec:AAC}

Consider a world that obeys the ordinary Laws of Physics.  The world $W$ has a set of possible states $S$, and its state $s[t]$ at time $t$ is some function of its state $s[t-1]$ at time $t-1$.  The world can thus be viewed as a form of dynamical system (probably non-linear, almost certainly stochastic) that evolves from state to state as time progresses.  These state transitions can be expressed as a \emph{transform} \ldots
\begin{equation} \label{eq:EQ1}
  f_{W}: s[t-1] \rightarrow s[t]    ,
\end{equation}
where $f_{W}$ is some function that transforms the state of the world at time $t-1$ to the state of the world at time $t$.

This means that the evolution of events in the world constitutes a continuous cycle of cause-and-effect.  Events follows a time course in which it can be said that \emph{actions} (i.e. the sequence of events in the past) lead to \emph{consequences} (i.e. the sequence of events in the future) which constitute further \emph{actions}, leading to further \emph{consequences}, and so on \ldots
\begin{equation} \label{eq:EQ2}
  Consequences = f_{W}\left(Actions \right)    .
\end{equation}

\begin{figure}[!h]
\centering
\includegraphics[width=12cm]{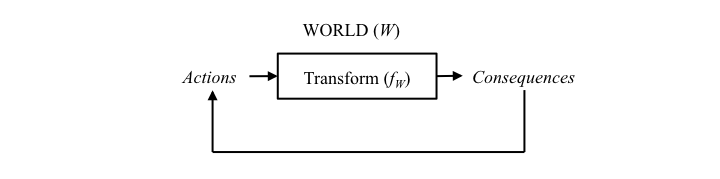}
\caption{Illustration of the continuous cycle of cause-and-effect in a world that obeys the ordinary Laws of Physics.}
\label{fig:AC}
\end{figure}

This straightforward scenario (as described by equations  \ref{eq:EQ1} and \ref{eq:EQ2}) can be expressed diagrammatically as shown in Fig.~\ref{fig:AC}.

Of course, while the transform $f_{W}$ might be relatively simple (since it is based on the Laws of Physics), the state-space $S$ of possible \emph{actions} and \emph{consequences} could be immense depending on the complexity of the world $W$.  This means that it is impossible to model everything that happens in the world.  However, in practice, some parts of the world might have very little influence on other parts.  So it is possible to consider a subset of the world $w$ that has a minimal dependency on the rest of the world $\bar w$.

\subsection{An Agent Manipulating the World} \label{sec:AMW}

\begin{figure}[!h]
\centering
\includegraphics[width=12cm]{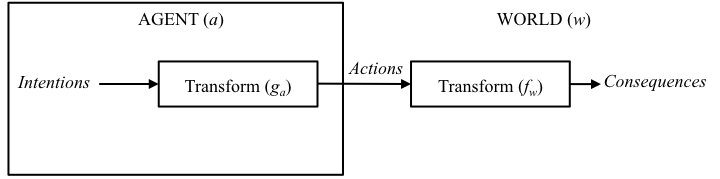}
\caption{Illustration of an intentional agent manipulating the world in an open-loop S-R configuration.}
\label{fig:AMW0}
\end{figure}

Now consider the presence of an intentional agent $a$ (natural or artificial) that seeks to effect a change in the world\footnote{The reason \emph{why} the agent wishes to change the state of the world is addressed later (in Section \ref{sec:NDCA}).}.  In this case the agent's intentions are converted into actions which are, in turn, transformed into consequences \ldots
\begin{equation}
  Consequences = f_{w}\left(g_{a} (Intentions) \right)    ,
\end{equation}
where $g$ is some function that transforms the agent's intentions into actions (a process known in robotics as `action selection' ) - see Fig.~\ref{fig:AMW0}.

\begin{figure}[!h]
\centering
\includegraphics[width=12cm]{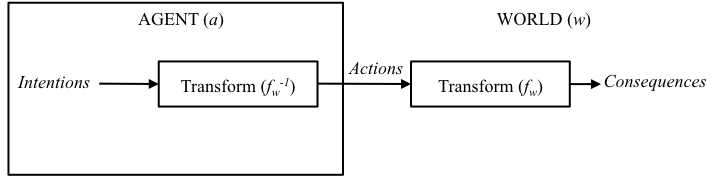}
\caption{Illustration of an intentional agent manipulating the world in an open-loop S-R configuration using an inverse model.}
\label{fig:AMW1}
\end{figure}

The situation illustrated in Fig.~\ref{fig:AMW0} corresponds to an open-loop S-R configuration, and this means that the accuracy with which an agent can achieve its intended consequences is critically dependent on having precise information about both $f$ and $g$.  Mathematically, the best method for achieving the required consequences is for the agent to employ an \emph{inverse transform} in which $g$ is replaced by $f^{-1}$ - see  Fig.~\ref{fig:AMW1}\footnote{Note that, in the situation where $f$ represents the plant operating a robot's actuators, the estimation of $f^{-1}$ is commonly referred to as `inverse kinematics' .}.

It is possible to discuss at length how information about the transforms $g$, $f$ or $f^{-1}$ could be acquired.  For example, model parameters could be computed using complex mathematical tools for system estimation or by employing machine learning techniques on extensive quantities of training data using a process known as `expectation maximisation' (EM) \cite{Do2008}.  Whichever approach is taken, the final outcome would inevitably be sensitive to any inaccuracies in calibrating the relevant model parameters as well as being unable to tolerate unforeseen noise and/or disturbances present in the agent or in the world.

\begin{figure}[!h]
\centering
\includegraphics[width=12cm]{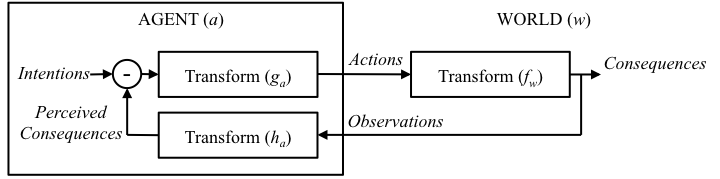}
\caption{Illustration of an intentional agent manipulating the world in a \emph{closed-loop} PCT-style negative-feedback configuration.  In this configuration the transform $g$ corresponds to the `output function' (in PCT) or `controller' (in classic control theory), and the transform $h$ corresponds to the `input function' (in PCT) or `feedback element' (in classic control theory).}
\label{fig:AMW1a}
\end{figure}

Control theory, and thereby PCT, of course provides an alternative \emph{closed-loop} solution that is not dependent on knowing $f$ (or $f^{-1}$).  An agent simply needs to be able to judge whether its objectives are being met - that is, whether the consequences of its actions \emph{match} its intentions.  An agent thus needs to be able to choose actions that minimise the difference between its intentions and the perceived consequences of its actions - see Fig.~\ref{fig:AMW1a}.  In PCT terminology, the agent's intentions correspond to the reference signal and the perceived consequences correspond to the controlled variable; actions are selected in order to minimise the difference between the two.

If the perceptual signal, and hence the error signal, is in the same parameter space as the control signal (see Fig.~\ref{fig:CCS}), then it is possible - in principle - to jump to the optimum solution in one step.  However, in practice it takes time to correct an error (since physical actions cannot take place instantaneously).  So the process typically \emph{iterates} towards a solution (by information flowing around the loop).  If, on the other hand, the perceptual signal and the control signal are not in the same parameter space, then there is no direct mapping between error and control action (unless such a mapping has been learnt in advance).  So, again, an iterative procedure is required\footnote{For example, a thermostat measures temperature, whereas the control signal for a heating/cooling system would be expressed in terms of power output.  This means that \emph{a-priori} there is no information on how to convert a difference between desired and actual temperature (the error signal) to an optimum power setting (the control signal).  Hence the power output has to be adjusted \emph{iteratively} until the desired temperature is reached.}.

This means that, although closed-loop control does not require information about $f$ (or $f^{-1}$), it does need to know about $g$ - the relationship between the error gradient in perceptual space and the appropriate control action.  Even a simple thermostat needs to be configured correctly such that a decrease in the temperature of a room leads to an increase in the level of heating (and \emph{vice versa}); in this case, the uni-dimensional perceptual error gradient determines the polarity of the action of the controller.  In general, perceptions and actions may lie in (different) high-dimensional spaces\footnote{In robotics, the space of possible actions is characterised by the Degrees-of-Freedom (DoF) in the system, and this usually corresponds to the number of joints or actuators/motors.} and, either the relationship between possible behaviours and the perceptual error gradient is (somehow) known in advance, or it has to be discovered (learnt) by active exploration; for example, using `reinforcement learning' \cite{Sutton1998} or the process referred to in PCT as `reorganisation' - see Section \ref{sec:HPCT}.

A further complication is that the error space may not be continuous.  In this case there is no gradient and hence no indication of which action to select in order to arrive at the correct solution.  For example, imagine a person who wishes to illuminate one section of a large room being faced with a set of unmarked light switches laid out in no obvious arrangement.  The only available strategy is to try each switch in a random fashion until the appropriate light illuminates\footnote{This is a common experience for an academic teaching in an unfamiliar lecture room.}.  Of course, once a mapping - $g$ - has been discovered, it may be stored in memory and used to guide future behaviour.

In many situations, negative feedback control is able to employ an optimisation technique known as \emph{gradient descent} in which the difference between the intentions (the reference signal) and the perceived consequences (the feedback signal) is a continuous variable that can be reduced monotonically to zero.  In this case, the dynamics of the optimisation is dependent on the appropriate choice of parameters in the controller (for example, $K_P$, $K_I$ and $K_D$ in a classic PID controller - as described in Section \ref{sec:CAC}); an inappropriate choice of controller parameters (such as the wrong sign for $K_P$) can even send the system in the opposite direction (that is, \emph{positive feedback}) with a consequent failure to converge.  In classic control systems, the process of optimising control parameters is often performed manually.  In PCT, it is performed by reorganisation (see Section \ref{sec:HPCT}) which is, itself, a negative feedback control process (and which is interestingly similar to the `actor-critic' approach to reinforcement learning \cite{Barto1983}).

This discussion makes it clear that, in general, \textbf{negative feedback control may be viewed as an iterative \emph{search}\footnote{Note that `search' is a term used by computer scientists to describe a process that engineers would refer to as `optimisation'.  In both cases, a system iterates/converges towards an optimum solution by minimising/maximising some `objective function' - in this case, minimising an error signal.  The value of using the term `search' is that it covers both continuous and discrete forms of optimisation.} process} and, in the example of the discrete light switches with no prior information, only a \emph{random} (as opposed to \emph{directed}) search is feasible.  The realisation that control can be regarded as a type of search is a key result, and it emphasises that the topology of the search space and the controller's information about that topology is key to the effectiveness and outcome of the optimisation.

\begin{figure}[!h]
\centering
\includegraphics[width=12cm]{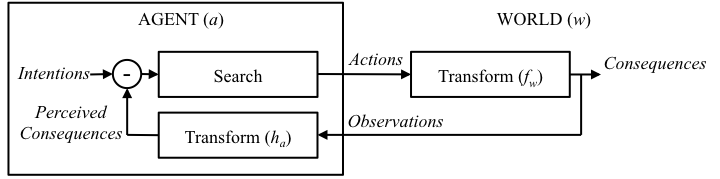}
\caption{Illustration of an intentional agent manipulating the world in a \emph{closed-loop} configuration using a search process to select actions that minimise the difference between its intentions and the perceived consequences.}
\label{fig:AMW2}
\end{figure}

Returning to the agent $a$ attempting to manipulate the world $w$, negative feedback control can thus be viewed as a search over possible actions to find those which give rise to the best match between intentions and perceived consequences.  This can be expressed as
\begin{equation}
  \widehat{Actions} = \underset{Actions}{\arg\min}\left(Intentions - Perceived\;Consequences \right)    ,
\end{equation}
where $ \widehat{Actions}$ represents an estimate of the actions required to minimise the difference between intentions and perceived consequences - see Fig.~\ref{fig:AMW2}.

The structure illustrated in Fig.~\ref{fig:AMW2} is similar to the classic PCT arrangement, but enhanced by the identification of iterative \emph{search} as the basis of a negative-feedback controlled optimisation process.  However, this configuration will only function correctly if (i) the agent can observe the consequences of its actions and (ii) the search space is concave\footnote{That is, the search space has only one \emph{global} optimum.}.  If the consequences of an agent's actions are hidden from observation in \emph{space} (because objects are obscured\footnote{For example, attempting to achieve some objective in the dark, reaching behind another object or, of particular significance here, manipulating the internal states of another agent (as will be discussed in Section \ref{sec:AMA}).}) or in \emph{time} (because feedback delays are too high or the intended consequences are in the future), the loop can still function, but only if the agent is able to \emph{estimate} the consequences of possible actions.  Likewise, if the search space is not concave but has many local minima, then an iterative search can avoid getting stuck in the nearest one by exploring the whole space \emph{in advance}.  In other words, in both of these cases, an agent could gain benefit by being able to \emph{predict} the consequences of possible actions.

\begin{figure}[!h]
\centering
\includegraphics[width=12cm]{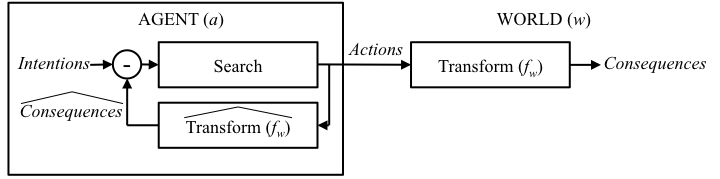}
\caption{Illustration of an intentional agent manipulating the world in a \emph{closed-loop} PCT-style configuration in the situation where it is unable to directly observe the consequences of its actions (in space or time).}
\label{fig:AMW3}
\end{figure}

In other words, if an agent cannot directly observe the consequences of its actions, or the search space has many local optima, then it needs to (i) estimate the relationship between possible actions and possible consequences ($f_w$), (ii) perform a search over hypothetical actions and then (iii) execute those actions that are found to minimise the estimated error.  In this case \ldots
\begin{equation}
  \widehat{Actions} = \underset{\widetilde{Actions}}{\arg\min}\left(Intentions - \widehat{f_{w}}(\widetilde{Actions}) \right)    ,
\end{equation}
where $\widehat{f_{w}}$ is the estimate of $f_w$ and $\widetilde{Actions}$ is the set of possible actions - see Fig.~\ref{fig:AMW3}.

What is interesting in this arrangement is that the estimated transform $\widehat{f_{w}}$ can be interpreted as a form of mental simulation (or predictor) which emulates the consequences of possible actions prior to action selection \cite{Grush2004}.  In other words, searching over $\widehat{f_{w}}(\widetilde{Actions})$ is equivalent to \emph{planning} (and corresponds to Powers' \emph{imagination mode} in HPCT, i.e. the \emph{b+c} setting in Fig.~\ref{fig:HPCTU}).  It also shows that the ability to perform ``what if'' simulations is not a property of a particular level of control, but is a generic property that can be instantiated in any control loop at any level (as proposed by Powers).

The ability to emulate the consequences of possible actions is important because, not only is it a link between PCT and AI planning, but it also offers the possibility of finding global (rather than local) solutions\footnote{As an example, the fastest route between two points on a map may not be the shortest, nor does each move forward necessarily take you nearer to the final destination.  This is the difference between local and global optimisation.}.  From an ecological perspective, this could be critical in terms of saving living systems vital time and/or energy in real world situations.

\subsection{An Agent Interpreting the World} \label{sec:AIW}

\begin{figure}[!h]
\centering
\includegraphics[width=12cm]{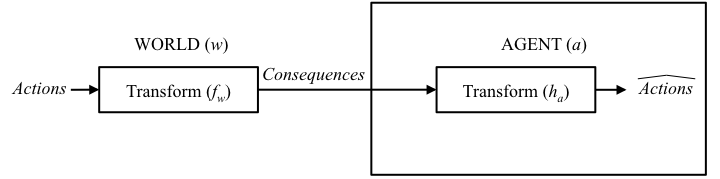}
\caption{Illustration of an agent $a$ attempting to infer the hidden causes/actions from their observable effects/consequences.}
\label{fig:AIW0}
\end{figure}

Having established an extended PCT-style framework for an agent attempting to manipulate the physical world (with search as the underlying mechanism supporting negative feedback control), it is now possible to turn to the complementary situation in which an agent $a$ is attempting to \emph{interpret} the world $w$.  In this case, interpretation is defined as an agent deriving potentially hidden actions/causes of events by observing their visible effects/consequences\footnote{For example, given an observation of an object suddenly accelerating along a flat surface, it is possible to \emph{infer} (using the laws of physics) that a hidden force must have acted upon that object.}.

\begin{equation}
  \widehat{Actions} = h_{a}\left(f_{w} (Actions) \right)    ,
\end{equation}
where $h$ is some perceptual function that transforms observed effects into estimated causes - see Fig.~\ref{fig:AIW0}.

\begin{figure}[!h]
\centering
\includegraphics[width=12cm]{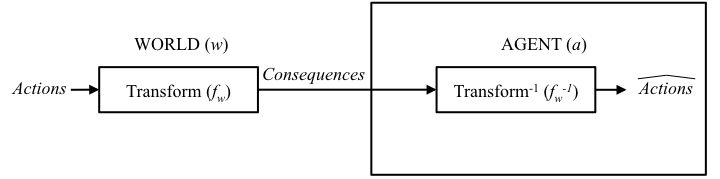}
\caption{Illustration of an agent $a$ attempting to infer the hidden causes/actions from their observable effects/consequences by means of the inverse transform $f_{w}^{-1}$.}
\label{fig:AIW1}
\end{figure}

Given that consequences are caused by actions via the transform $f_w$, it is possible (in principle) to compute the actions directly from the observed consequences using the inverse transform $f_{w}^{-1}$ - see Fig.~\ref{fig:AIW1}.

Of course, the accuracy of this process depends on the fidelity of the inverse transform.  In practice, $f_{w}^{-1}$ is not known and very hard to estimate.  A more tractable solution is thus not to use an inverse model, but to construct a so-called `forward model'\footnote{Also known as a `generative model' .} (that is, an estimate of $f_w$) and to compare its output with the observed signals.  Mathematically, this is the optimum way to estimate hidden variables\footnote{A `hidden variable' is a measurement which cannot be made directly (e.g. by observation).} given uncertainty in both the observations and the underlying process.  Such a configuration is known as a `maximum likelihood (or \emph{Bayesian}) classifier'.  Also, it is a standard result in the field of statistical estimation that the parameters of forward/generative models are typically much easier to derive using reliable techniques for maximum likelihood (ML) or maximum \emph{a-posteriori} (MAP) estimation, and performance degrades gracefully when there is missing data.  

The process of interpretation using such an arrangement thus proceeds by searching over possible actions/causes to find the best match between the predicted and the observed consequences \ldots
\begin{equation}
  \widehat{Actions} = \underset{Actions}{\arg\min}\left(Consequences - \widehat{f_w}(Actions) \right)    .
\end{equation}

\begin{figure}[!h]
\centering
\includegraphics[width=12cm]{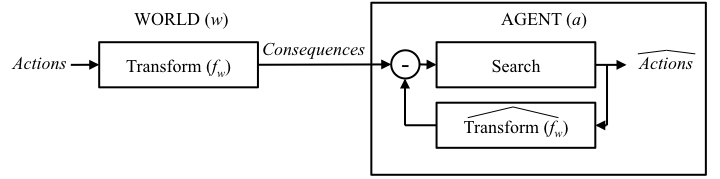}
\caption{Illustration of an agent $a$ attempting to infer the hidden causes/actions from their observable effects/consequences by using a negative-feedback control loop to search the outputs from $\widehat{f_w}$ (a forward estimate of $f_w$).}
\label{fig:AIW2}
\end{figure}

This process is illustrated in Fig.~\ref{fig:AIW2}, and what is immediately apparent is that, just as the manipulation case, the process of \textbf{interpretation may also be construed as a negative-feedback control loop} (in this case, a `search' over possible explanations).  This is a significant outcome that is not explicit in (H)PCT; hence it provides a potentially valuable link between PCT and mainstream approaches to machine perception.

In fact, the architecture illustrated in Fig.~\ref{fig:AIW2} is a familiar\footnote{\ldots in the field of machine learning.} model-based \emph{recognition} framework in which the recognition/interpretation/inference of the (hidden) cause of observed behaviour is viewed as a search over possible outputs from a forward model that is capable of generating that behaviour \cite{Wilson2005,Pickering2013}.  This is an established and powerful approach in machine perception and scene analysis (known as \emph{analysis-by-synthesis}\footnote{The term `analysis-by-synthesis' refers to an established process in the field of signal processing in which the value of an unknown variable is estimated from observation data (`analysis') using a model of how the observations might have been generated (`synthesis').}), and it has been applied very successfully in a range of practical scenarios \cite{Chin1986,Weiss2001,Demiris2005}.

\subsection{An Agent Communicating its Intentions to Another Agent} \label{sec:AMA}

The foregoing establishes a remarkably symmetric framework for manipulating and interpreting the world in the presence of uncertainty and unknown disturbances; both employ PCT-style negative-feedback control loops that perform a search over the potential outputs of forward models.  This section (and the next) extends the arguments to the case where the world contains more than one agent: a world in which a sending agent $s$ is attempting to change the mental state of a receiving agent $r$ (that is, \emph{communicating} its intentions without being able to directly observe whether those intentions have been perceived\footnote{It might seem perverse to rule out feedback.  However, the aim here is to consider the general case, where feedback may be present or absent.}) and the receiving agent is attempting to interpret the sending agent (that is, estimating the sending agent's intentions based only on observations of its actions).

\begin{figure}[!h]
\centering
\includegraphics[width=12cm]{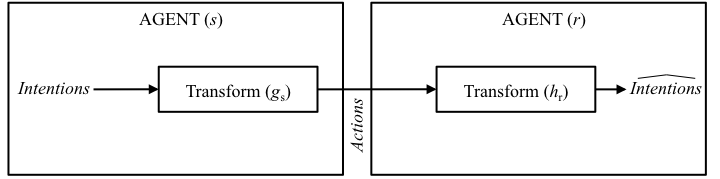}
\caption{Illustration of a world containing two communicating agents - a sender $s$ and a receiver $r$.}
\label{fig:SR}
\end{figure}

So, again starting from first principles - for the sending agent $s$
\begin{equation}
  Actions_s = g_s \left(Intentions_s \right)    ,
\end{equation}
where $g_s$ is the transform from intentions to behaviour, and for the receiving agent $r$
\begin{equation}
  Interpretations_r = h_r \left(Actions_s \right)    ,
\end{equation}
where $h_r$ is the transform from observed behaviour to interpretations - see Fig.~\ref{fig:SR}.

\begin{figure}[!h]
\centering
\includegraphics[width=12cm]{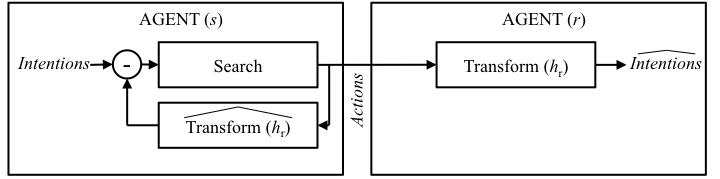}
\caption{Illustration of a sending agent optimising its communicative behaviour by inferring the interpretations made by the receiving agent using a process of \emph{synthesis-by-analysis}.}
\label{fig:SA}
\end{figure}

Hence, for agent $s$ attempting to communicate its intentions to agent $r$, the arguments put forward in Section \ref{sec:AMW} suggest that, if there is no direct feedback from agent $r$, then agent $s$ needs to compute appropriate behaviour (actions) based on
\begin{equation}
  \widehat{Actions_s} = \underset{\widetilde{Actions_s}}{\arg\min}\left(Intentions_s - \widehat{h_r}(\widetilde{Actions_s}) \right)    ,
\end{equation}
which is a negative-feedback control loop performing a search over possible behaviours by agent $s$ and their interpretations by agent $r$ as estimated by agent $s$ - see Fig.~\ref{fig:SA}.  This process can be viewed as \emph{synthesis-by-analysis}.

\subsection{An Agent Interpreting the Behaviour of Another Agent} \label{sec:AIA}

\begin{figure}[!h]
\centering
\includegraphics[width=12cm]{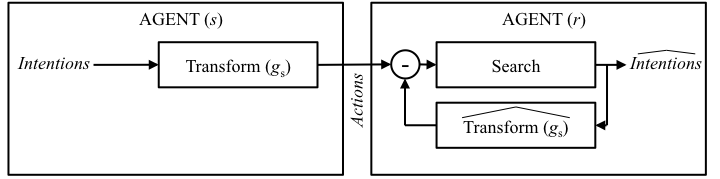}
\caption{Illustration of a receiving agent inferring the communicative intentions of a sending agent using a process of \emph{analysis-by-synthesis}.}
\label{fig:RA}
\end{figure}

For agent $r$ attempting to interpret the intentions of agent $s$, the arguments put forward in Section \ref{sec:AIW} suggest that agent $r$ needs to compare the observed actions of agent $s$ with the output of a forward model for agent $s$
\begin{equation}
  \widehat{Intentions_s} = \underset{Intentions_s}{\arg\min}\left(Actions_s - \widehat{g_s}(Intentions_s) \right)    .
\end{equation}
which is a negative-feedback control loop performing a search over the possible intentions of agent $s$ and their realisations by agent $s$ as estimated by agent $r$ - see Fig.~\ref{fig:RA}.  As in Fig.~\ref{fig:AIW2}, this process can be viewed as \emph{analysis-by-synthesis}.

Interestingly, this configuration is exactly how contemporary approaches to automatic speech recognition are formulated (using a probabilistic forward generative model known as a `Hidden Markov Model' - HMM) \cite{Gales2007}).  In fact, the analysis-by-synthesis approach to speech recognition is not only reminiscent of the \emph{Motor Theory} of speech perception \cite{Liberman1967}, but is also supported by recent neuroimaging data \cite{Skipper2014,Kuhl2014}.  Hence, there is ample evidence that such a configuration is appropriate for modelling perceptual inference.

\subsection{Using \emph{Self} to Model \emph{Other}} \label{sec:SMO}

\begin{figure}[!h]
\centering
\includegraphics[width=12cm]{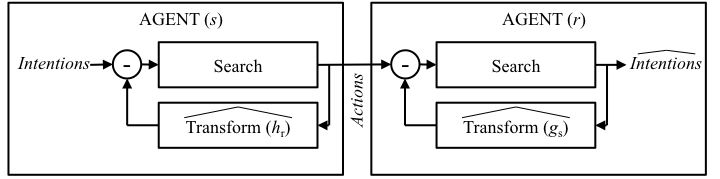}
\caption{Illustration of a world containing one agent (a sender $s$) communicating with another (a receiver $r$) where each makes use of a model of the other.  The sender infers the interpretation of its intentions by the receiver using \emph{synthesis-by-analysis}, and the receiver infers the intentions of the sender using \emph{analysis-by-synthesis}.}
\label{fig:SR2}
\end{figure}

Looking at the arrangements outlined in Sections \ref{sec:AMA} and \ref{sec:AIA}, we arrive at an important result; both require one agent to have a model of (some aspect of) the other.  The sending agent $s$ selects its actions by searching over possible interpretations by the receiving agent $r$ using an estimate of the receiving agent's transform from observations to interpretation ($\widehat{h_r}$).  The receiving agent $r$ infers the intentions of the sending agent $s$ by searching over possible interpretations using as estimate of the sending agent's transform from intentions to actions ($\widehat{g_s}$) - see Fig.~\ref{fig:SR2}.  

\begin{figure}[!h]
\centering
\includegraphics[width=12cm]{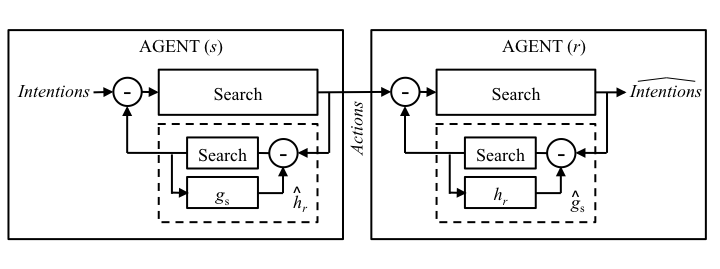}
\caption{Illustration of sending and receiving agents exploiting knowledge of themselves in order to communicate.}
\label{fig:ASAS}
\end{figure}

The configuration shown in Fig.~\ref{fig:SR2} leads to an interesting question: where do the transforms $\widehat{h_r}$ and $\widehat{g_s}$ come from?  More precisely, how might their parameters be estimated?  Obviously they could be derived using a variety of different learning procedures (including PCT-style reorganisation).  However, one intriguing possibility is that, because of the similarity between agents\footnote{This situation is particularly relevant to the high degree of similarity that is found in living systems between conspecifics (members of the same species).}, \textbf{each agent could approximate these functions using information recruited from their own structures}\footnote{The idea of recruiting information about \emph{self} in order to model \emph{other} is a core component of the PRESENCE architecture (Section \ref{sec:TWF}) which, in turn, is founded on the principle of mirror structures in the brain (Section \ref{sec:ACS}).}.  In other words, $\widehat{h_r} \mapsfrom h_s$ (which can be searched using $g_s$ rather than $\widehat{g_r}$) and $\widehat{g_s} \mapsfrom g_r$ (which can be searched using $h_r$ rather than $\widehat{h_s}$) - see Fig.~\ref{fig:ASAS}.   

\begin{figure}[!h]
\centering
\includegraphics[width=12cm]{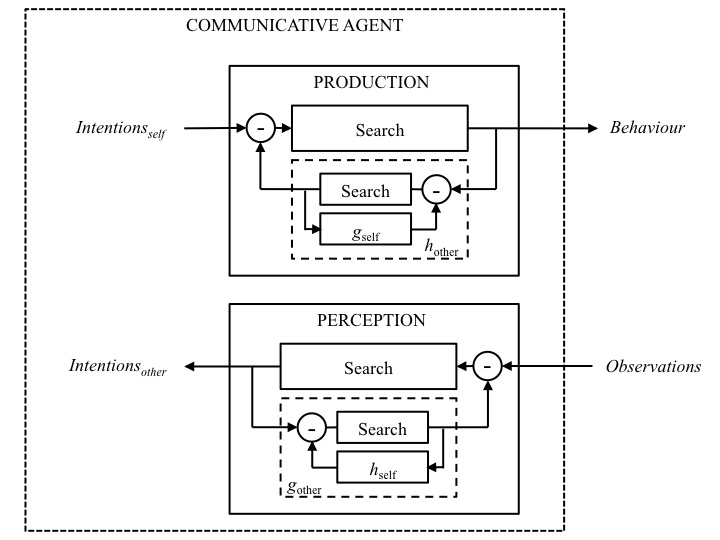}
\caption{Illustration of a communicative agent that is capable of optimising the signalling of its own intentions (PRODUCTION) and inferring the intentions of others (PERCEPTION).}
\label{fig:PRODPERC}
\end{figure}

This arrangement, in which both agents exploit sensorimotor knowledge of themselves to model each other, can be thought of as \emph{synthesis-by-analysis-by-synthesis} for the sending agent and \emph{analysis-by-synthesis-by-analysis} for the receiving agent.  Combining both into a single communicative agent gives rise to a structure where perception and production are construed as parallel recursive control feedback processes (both of which employ search as the underlying mechanism for optimisation), and in which the intentions of self and the intentions of other are linked to the behaviour of self and the observations of other, respectively - see Fig.~\ref{fig:PRODPERC}.

\subsection{A Needs-Driven Communicative Agent} \label{sec:NDCA}

The preceding arguments have provided interesting answers to two key questions: (i) how can an agent optimise its behaviour in order to to communicate its intentions and (ii) how can an agent infer the intentions of another agent by observing their behaviour?  However, thus far it has been assumed that intentionality is a key driver of communicative interaction.  Hence, an obvious question is - where do the intentions come from?  Of course, HPCT provides a clear answer; the reference signals at one level in a hierarchy are set by the level above and mediated by memory (as illustrated in Fig.~\ref{fig:HPCTU}).  So this would suggest that intentions should be the output from some higher level.

In B:CP \cite{Powers1973}, Powers doesn't use the term `intention' as such, rather he talks about purposeful behaviour (at \emph{all} levels of the HPCT hierarchy).  However, by invoking intentionality as a manifestation of purposeful goal-driven behaviour, it is possible to make a direct link between HPCT and contemporary architectures such as DAC (Section \ref{sec:ACS}) and BDI (Section \ref{sec:ABM})\footnote{The main difference between HPCT and these architectures (including the one developed here) is that HPCT provides a rich decomposition of purposeful behaviour into the necessary levels of detailed control, whereas the other schemes tend to collapse such levels into a single abstraction.}.  In particular, the BDI approach to agent-based modelling makes it clear that intentionality is not only derived from an agent's longer-term goals and desires, but that it is also conditioned by the beliefs that an agent holds.  More directly, the DAC architecture emphasises that behaviours are ultimately driven by a motivational system based on an agent's fundamental \emph{needs} (such as energy for survival).  This appears to be similar to the important role that intrinsic control systems play in the development of the perceptual hierarchy in HPCT \cite{Powers1973}.

\begin{sidewaysfigure}
\centering
\includegraphics[width=19cm]{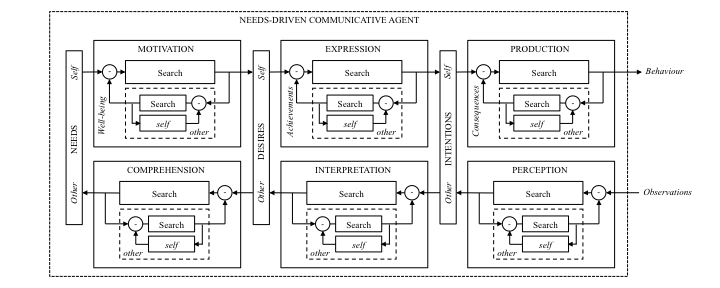}
\caption{Illustration of the proposed architecture for a needs-driven communicative agent.}
\label{fig:NDCA}
\end{sidewaysfigure}

Putting all this together, it is possible to formulate a generic (and remarkably symmetric) architecture for a \emph{needs-driven} communicative agent that is both a sender and a receiver\footnote{A forerunner of this framework has been referred to as MBDIAC (Mutual Beliefs Desires Intentions Actions \& Consequences) \cite{Moore2014a}.} - see Fig.~\ref{fig:NDCA}.  In this framework it is proposed that a communicative agent's behaviour is conditioned on appropriate motivational and deliberative states (reference signals): \emph{Needs} \textrightarrow\;\emph{Desires} \textrightarrow\;\emph{Intentions}.  Likewise, the intentions, desires and needs of another agent are inferred via a parallel interpretive structure: \emph{Perception} \textrightarrow\;\emph{Interpretation} \textrightarrow\;\emph{Comprehension}\footnote{From a PCT perspective, these are all perceptions.}.

\section{Discussion} \label{sec:DISC}

The foregoing section has established a putative framework for a needs-driven communicative agent in which the data structures for needs, desires and intentions represent the informational \emph{belief} states (controlled variables) of the agent.  As such, these representational structures are analogous to the memory layers in HPCT (as shown in Fig.~\ref{fig:HPCTU}).  However, a key feature of the configuration proposed here (for a communicative agent) is that these structures explicitly contain information relating to both self and other, and this allows the behaviour of self to be conditioned on the inferred informational state of other.  In other words, it provides a basis for modelling interpersonal stances such as \emph{empathy} \cite{Iacoboni2005} and \emph{Theory of Mind} (ToM) \cite{Premack1978}\footnote{Interestingly, Marken - a long-standing PCT practitioner - has recently posited the value of PCT in testing the validity of the inferences that observers make about the intentional states of others \cite{Marken2013}.  This might be usefully carried over into the proposed framework as a strategy an agent might employ in order to establish a desired cooperative/competitive relationship with another agent.}.

Another interesting aspect of the architecture illustrated in Fig.~\ref{fig:NDCA} is that, like DAC and PRESENCE, it is founded on a motivational system based on \emph{needs} (for example, as described by Maslow in his famous `Hierarchy of Needs' \cite{Maslow1943}).  This might seem to be an unnecessary embellishment - a feature that is more relevant to modelling a living system than to designing an artificial agent.  However, invoking such a needs-based framework answers a fundamental question - why would an agent (natural or artificial) do anything \cite{Oudeyer2007}?  Also, it is envisaged that it would be useful to partition needs into those that pertain to physical health (such as survival and safety needs) and those that pertain to psychological health (such as personal and social needs), with the relevant feedback signals being related to physiological and mental well-being.  This not only ties in closely with Powers' observations concerning the importance of optimising \emph{intrinsic} variables that are related to the health of the organism, but it also links with his view that the control of such variables provides the basis for reorganisation (i.e. structural learning) \cite{Powers1973}.

In practice, the motivational framework can be decomposed into two elements: (i) a needs structure that provides the incentives for behaviour and (ii) the amount of effort that will be devoted to meeting those needs.  The previous paragraph (and the architecture shown in Fig.~\ref{fig:NDCA}) addresses the first of these. The second relates to the \emph{enthusiasm} of an agent, i.e. how much it cares about meeting its needs.  In a conventional PCT-style negative-feedback control loop, such behaviour would correspond to the loop gain; a high loop gain implying a high degree of physical effort/enthusiasm and \emph{vice versa}.  In the framework outlined here, such behaviour would relate to the depth and quality of the search(es) that pervade the overall structure - deep search(es) implying a high degree of mental effort/enthusiasm and \emph{vice versa}.

Interestingly, the proposed framework also provides a practical architecture that supports appraisal theories of emotion \cite{Scherer2001,Marsella2010} based on the classic dimensions of pleasure and arousal \cite{Mehrabian1996}.  For example, the comparators that pervade the structure each constitute mini-appraisal units that generate measures of positive/negative affect - the error signals \cite{Carver1998}.  Likewise, the effort that an agent devotes to achieving its goals and intentions (as discussed in the previous paragraph) can be be interpreted as a measure of arousal.  Tying this in with Powers' notion of intrinsic variables driving reorganisation, these emergent properties of the proposed framework can be interpreted as (i) emotion driving behaviour and (ii) the perception of emotion driving learning\footnote{This overlaps nicely with Powers' description of the role of emotion \cite{Powers1973}.}.

Finally, the core components in the proposed framework have been portrayed in Fig.~\ref{fig:NDCA} as a three-layered control structure.  However, this is just a necessary simplification in order to establish the basic principles.  Also, the emphasis has been on a communicative agent that seeks to influence the internal states of another agent (as opposed to an agent that seeks to manipulate the physical behaviour of another agent).  In practice, the structures outlined here would be decomposed into a richer hierarchy of parallel control loops - much as envisaged in HPCT.  The main difference would be that the perceptual apparatus in the proposed framework would be more structured than in HPCT, with explicit parameter sharing between behavioural and perceptual components (rather than the switching arrangement illustrated in Fig.~\ref{fig:HPCTU}).  Also, whilst Powers' mechanism for reorganisation should be capable of optimising the structure of a decomposed architecture for any given problem, in practice no-one has yet succeeded in developing a practical solution.  Since some of the solutions presented here are related to more contemporary approaches to machine learning, it is envisaged that they might offer a way forward in understanding how to configure an HPCT structure automatically.

\section{Summary and Conclusion} \label{sec:CONC}

This chapter has shown how a practical framework for modelling and implementing `intelligent' systems can be developed from the establishment of a set of fundamental principles which both intersects with and extends Perceptual Control Theory.  It has been argued that PCT has hitherto placed less emphasis on the transformation of perceptual functions through mental simulation, and has treated the mapping from sensation to perception as a relatively straightforward series of transformations (albeit mediated by memory), thereby overlooking the potential of powerful generative model-based solutions that have emerged in practical fields such as visual or auditory scene analysis.  Starting from first principles, and considering how an intelligent agent might interact with the world and with other agents it might contain, it has not only been shown how these ideas might be integrated into PCT, but also how PCT might be extended towards a remarkably symmetric architecture for a needs-driven communicative agent.

The arguments presented in Section \ref{sec:TIS} first led to the realisation that the optimisation which takes place within a negative feedback control loop is equivalent to a search process, and this led to a reformulation of intentional behaviour as a search over possible actions in order to find those which minimise the difference between an agent's intentions and the perceived consequences.  These principles were then extended to cover the situation in which the consequences of actions might be hidden from direct observation (for example, when one agent is manipulating the mental state of another agent), and this led to the emergence of simulation as a key mechanism for maintaining a control loop structure by using an estimate/prediction of possible action consequences.  It was then revealed how searching a simulation facilitates off-line planning which has the important benefit of being able to compute global (rather than local) solutions.

These principles were then extended to an agent attempting to interpret behaviour, and it was shown how the perceptual process itself can be configured using a familiar PCT-style negative feedback control loop (also involving simulation) which turns out to be equivalent to established techniques for model-based recognition.  It was then shown how the arguments could be extended to cover agent-to-agent communication, and this led to the crucial realisation that the parameters of the models used by one agent for simulating another agent could be derived from the models used to manage their own behaviour - that is, using \emph{self} to model \emph{other}.  This novel arrangement was then embedded within a BDI (beliefs, desires, intentions) structure to create a new framework for a needs-based communicative agent.

One of the encouraging outcomes of this analysis is that the new perspective not only relies on a central role being played by PCT-style negative-feedback control in action selection \emph{and} interpretation, but it also links with recent discoveries in cognitive neuroscience (such as the sensorimotor overlap attributed to mirror neurons) and aligns in interesting ways with neurally-inspired architectures such as DAC (Section \ref{sec:ACS}).  Also, since it has been shown that negative feedback control (as an optimisation process) may be viewed as a type of search, this opens up a potentially important link with a range of model-based techniques that are already established in some areas of contemporary AI and Cognitive Systems (particularly in the field of spoken language processing).  This means that the development of the approach can benefit from progress being made in these other fields, as well as extending the scope and influence of PCT to provide a more parsimonious computational framework for `intelligent' systems (whether natural or artificial).

Based on the framework established thus far, the critical next steps are to formulate practical mechanisms for learning local/global structures and to investigate the implications of simulations of self and other using small scale agents (such as robots) in tasks which involve real-world interaction and communication with other agents.  It is too soon to extend the principles to complex scenarios (such as full blown language-based interaction).  Rather, it is necessary to scale-up the complexity in a careful and controlled manner in order to maintain a firm scientific and mathematical foundation for the entire enterprise.  Such work is already underway, and early results are encouraging.

Finally, the arguments presented herein lead to the following overall conclusion with respect to modelling intelligent communicative agents - if behaviour is the control of perception (the central tenet of Perceptual Control Theory), then perception (at least for communicative agents) can be said to be the \emph{simulation} of behaviour.

\section{Acknowledgements}

The author would like to thank colleagues in the Sheffield Centre for Robotics (SCentRo) and the Bristol Robotics Laboratory (BRL) for discussions relating to the content of this chapter.   This work was partially supported by the European Commission [grant numbers EU-FP6-507422, EU-FP6-034434, EU-FP7-231868, FP7-ICT-2013-10-611971] and the UK Engineering and Physical Sciences Research Council [grant number EP/I013512/1].

\end{document}